\documentclass{article}

\usepackage{arxiv}

\usepackage[utf8]{inputenc}
\usepackage{CJKutf8} 
\usepackage[T1]{fontenc}    
\usepackage{hyperref}       
\usepackage{url}            
\usepackage{booktabs}       
\usepackage{amsfonts}       
\usepackage{nicefrac}       
\usepackage{microtype}      
\usepackage{graphicx}
\usepackage{natbib}
\usepackage{doi}
\usepackage{amsmath}
\usepackage{enumitem}

\title{Soul Computing: A Theoretical Framework and Technical Architecture for Intelligent Agents with Independent Consciousness}

\author{%
	\href{https://orcid.org/0000-0003-3427-9014}{\includegraphics[scale=0.06]{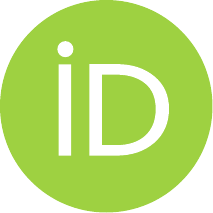}\hspace{1mm}Jinshan Zhang} \\
	Innovation and Management Center\\
    School of Software Technology\\
	Zhejiang University\\
	(Ningbo), Ningbo, China \\
	\texttt{zhangjinshan@zju.edu.cn} \\
	\And
	\href{https://orcid.org/0009-0008-6497-9322}{\includegraphics[scale=0.06]{orcid.pdf}\hspace{1mm}Xishi Zhou} \\
	School of Software Technology\\
	Zhejiang University\\
	Ningbo, China \\
	\texttt{zhouxishi@zju.edu.cn} \\
	\And
	\href{https://orcid.org/0009-0001-2147-792X}{\includegraphics[scale=0.06]{orcid.pdf}\hspace{1mm}Qiu Peng} \\
	School of Software Technology\\
	Zhejiang University\\
	Ningbo, China \\
	\texttt{PockeyQ@zju.edu.cn} \\
	\And
	\href{https://orcid.org/0000-0003-4703-7348}{\includegraphics[scale=0.06]{orcid.pdf}\hspace{1mm}Jianwei Yin} \\
	School of Software Technology\\
	Zhejiang University\\
	Ningbo, China \\
	\texttt{zjuyjw@cs.zju.edu.cn} \\
}

\renewcommand{\shorttitle}{Soul Computing}

\hypersetup{
pdftitle={Soul Computing: A Theoretical Framework and Technical Architecture for Intelligent Agents with Independent Consciousness},
pdfsubject={cs.AI, cs.CY},
pdfauthor={Jinshan Zhang, Xishi Zhou, Qiu Peng, Jianwei Yin},
pdfkeywords={Soul Computing, Digital Consciousness, Independent Agent, Affective Computing, Mortal Computation},
}

\begin{document}
\maketitle

\begin{abstract}
Breakthroughs in large language models and multimodal generation technologies have propelled the digital reconstruction of human mental traits, emotional patterns, and long-term memory from science fiction toward engineering practice. Yet current research and industry practices at the intersection of AI and digital humans remain hampered by fundamental conceptual ambiguities: the essential differences between next-generation intelligent agents and traditional virtual humans, the construction pathways for digital entities possessing self-identity, and the core technical and ethical challenges confronting this domain all demand urgent clarification. This paper systematically examines the transformative logic underlying the transition from traditional virtual humans to the ``Soul Computing'' paradigm, driven by frontier AI technologies. We first analyze the evolutionary patterns of human consciousness and memory mechanisms, reassessing the core value of massive multimodal digital fragments in the reverse reconstruction of individual mental worlds. On this basis, we formally delineate the academic connotations of narrow and broad Soul Computing for the first time, clarifying its academic boundaries and essential distinctions from Affective Computing, Historical Reconstruction, and Mortal Computation. We argue that Soul Computing systems must architecturally construct an ``Intensional'' core rather than serving as purely ``Extensional'' functional carriers, thereby enabling the fundamental transition of AI from toolhood to living agency. We further propose a hierarchical technical architecture encompassing a data-driven layer, a narrow computing core layer, and a broad externalization layer, and identify five core challenges that must be overcome, providing a forward-looking theoretical framework and practical guidance for research and industrial deployment of next-generation intelligent agents with independent consciousness.
\end{abstract}


\section{Introduction}
\label{sec:introduction}

Over the half-century since the birth of computer science, humanity's use of computational means to transform the objective world has undergone several fundamental paradigm shifts. In early computer-aided design, the academic community consistently sought to fuse human creativity with the high-performance computational capacity of machines---a vision that directly propelled design technology from two-dimensional drafting to three-dimensional solid modeling and feature-based parametric design. Yet regardless of iterative advances, traditional computing systems never escaped the logic of extensional instrumentality: humans served as the originators of creativity and intent, defining unstructured requirements and making core decisions, while computers performed only highly structured mathematical analysis and mechanical execution as tools.

As early as the 1990s, Qian Xuesen presciently transcended the prevailing global view of virtual reality as merely a visualization and interaction tool, rendering it instead as ``Lingjing'' \begin{CJK}{UTF8}{gbsn}(灵境)\end{CJK}---a term imbued with Eastern philosophical resonance---and argued that the ultimate value of this technology lay in realizing intelligent leap through human-machine fusion, building a bridge for the mutual sustenance and coexistence between the mental world of carbon-based life and silicon-based digital space. Qian's vision of ``Lingjing'' extended far beyond scene rendering and interaction simulation; it pointed directly to the core proposition of the subsistence and extension of human consciousness, memory, and spiritual essence in the digital world, laying a theoretical foundation of both scientific foresight and humanistic depth for decades of subsequent research. Now, in an era of explosive advancement in AI, humanity is attempting through foundational algorithmic breakthroughs to answer this proposition spanning over three decades---to confront the more ultimate question: how can the spiritual core, long-term memory, and autonomous consciousness of an individual life---quantities that resist measurement---be faithfully replicated and preserved in silicon-based media?

With digital existence now a social norm, human society is experiencing an unprecedented explosion of data across all domains \citep{negroponte1995being}. Individuals leave massive and highly fragmented digital traces across social media, instant messaging, collaborative work platforms, and sensor networks pervading physical space---encompassing text logs, voice messages, video recordings, and fine-grained behavioral timestamps and spatial trajectories. Traditional data mining has largely treated these digital traces as static commercial assets, focusing on extracting marketing value and constructing macro-level demographic profiles, while broadly neglecting their deeper logical value as constituent pieces of individual consciousness. These behavioral footprints scattered across digital space are not merely carriers of objective information; they constitute genuine physical mappings of an individual's unique personality traits, cognitive habits, value preferences, and emotional fluctuation patterns.

The leapfrog development of large language models and multimodal generation technologies in recent years has provided unprecedented technical support for assembling these digital fragments and reconstructing individual mental worlds across the boundaries of life \citep{openai2023gpt4v, radford2021learning}. It is at this historic convergence of technological evolution and theoretical demand that ``Soul Computing''---an interdisciplinary frontier concept---has emerged. Its core proposition: if deep learning algorithms can achieve deep semantic alignment and logical recombination of these fragmented, unstructured, noise-laden digital footprints, can a digital entity possessing vitality, continuous memory streams, and real-time emotional feedback be reconstructed within a silicon-based architecture?

This proposition not only surpasses the underlying technology of existing static digital humans and script-driven dialogue systems, but redefines the form of life at the philosophical level, resonating across time with Qian Xuesen's prescient vision of human-machine symbiosis and spiritual perpetuation in his ``Lingjing'' concept. Turing Award laureate Geoffrey Hinton has articulated the essential distinction between ``Mortal Computation'' and ``Immortal Computation'' \citep{hinton2023two}. He observes that intelligent realization in carbon-based brains depends on tight physical coupling between software and hardware; the biological demise of an individual inevitably entails the permanent annihilation of its specific knowledge, memory, and consciousness. In contrast, silicon-based digital computing achieves complete decoupling of software (weight matrices bearing knowledge and consciousness) from hardware (physical chips and storage media), making the theoretical possibility of ``immortality'' at the consciousness level achievable through algorithmic training and lossless transfer of weight parameters. Soul Computing constitutes the core technical pathway for engineering this grand philosophical vision.

Concurrently, industry has taken the lead in frontier exploration and technical deployment in this domain. Meta Platforms filed and was granted a patent between 2023 and 2024 for a system based on large language models that simulates the social media behavior of deceased users (US Patent 12513102B2) \citep{meta2024techniques}. The system described therein can, through deep analysis of a user's posts, comment interactions, like preferences, and private communications, continue posting content and interacting with friends and family in the user's distinctive linguistic style, value orientation, and interaction logic after the user's death or prolonged disengagement from online environments, while also possessing the technical potential to generate highly realistic audio-visual content. Despite facing public ethical controversy, Meta stated it has no immediate plans to productize the patent at scale; yet the disclosure of its underlying technical architecture has clearly validated the substantial feasibility and application prospects of Soul Computing in scenarios such as digital legacy inheritance, cross-temporal emotional social interaction, and cyborg symbiosis.

As with all intelligent technologies in their early stages, however, the field of Soul Computing remains in the early phase of theoretical construction, plagued by conceptual conflation and ambiguous positioning. Numerous research institutions and commercial teams frequently conflate it with traditional Affective Computing \citep{pei2024affective, picard1997affective}, retrieval-augmented generation (RAG)-based historical figure Q\&A systems \citep{xu2024character}, and Mortal Computation \citep{ororbia2023mortal}. To clarify the academic lineage and define the direction of technical evolution, this paper aims to systematically construct the theoretical paradigm and technical architecture of Soul Computing. The subsequent sections are organized as follows: we begin from the evolutionary patterns of human consciousness and memory mechanisms to elucidate the core role of digital fragments in reconstructing individual mental worlds, establishing the technical boundaries and core mechanisms of Soul Computing; we then systematically review the literature to clarify its essential distinctions from Affective Computing, Historical Reconstruction, and Mortal Computation; finally, we propose a forward-looking hierarchical technical architecture for Soul Computing and identify five core challenges at the data, algorithmic, and ethical levels that demand breakthroughs.

\section{The Consciousness Formation Process and Ontological Mapping of Digital Fragments}
\label{sec:consciousness}

To clarify the core role that AI should assume within the Soul Computing paradigm and its implementation pathways, one must first systematically deconstruct the natural construction mechanisms of human consciousness and personality traits, and then establish the ontological mapping value of individual digital footprints in the reverse reconstruction of consciousness. Just as a deep analysis of the cognitive processes in engineering design is a prerequisite for defining the functional boundaries of next-generation intelligent design systems, a systematic understanding of the Life and Consciousness Process is the theoretical cornerstone for delineating the underlying algorithmic logic and establishing the technical implementation pathways of Soul Computing.

\subsection{A Dynamic Evolutionary Cognitive Model of Individual Consciousness}
\label{sec:consciousness-dynamic}

Research in modern cognitive psychology and neurobiology demonstrates that an individual's personality traits and independent consciousness are neither innately pre-encoded static programs nor fixed behavioral scripts. Rather, they are dynamic systems gradually constructed through the continuous interaction of the brain with the external environment, via long-term deposition of memory, dynamic emotional feedback, and iterative behavioral refinement. This highly complex consciousness construction process can be abstracted at the macro level into four mutually nested, continuously coupled core evolutionary phases---none of which are linearly sequential isolated steps, but rather dynamic cyclic processes spanning an individual's entire life cycle.

The first phase is the perception and feature encoding of external information. Individuals continuously receive massive informational stimuli from the external environment through physiological senses; the nervous system performs selective filtering, hierarchical encoding, and feature extraction on input signals, ultimately forming primary perceptual representations of specific events, persons, and environments \citep{hubel1962receptive}. This phase constitutes the starting point for an individual's connection with the external world and the informational input foundation for all higher-order cognitive activity.

The second phase is the hierarchical deposition of episodic and semantic memory. The hippocampus does not replicate external information in full fidelity; rather, it selectively encodes high-frequency cognitive patterns or experiences accompanied by strong emotional valence (such as profound joy or deep fear) into long-term episodic memory \citep{tulving1972episodic, squire1991medial}. Over time, universal patterns and cognitions within episodic memory are further abstracted and distilled into semantic memory---an individual's understanding of the rules governing how the world operates---and the continuous construction of semantic memory ultimately establishes the cognitive foundation for subjective value judgments about specific matters \citep{tulving1985memory}.

The third phase is the consolidation of personalized thinking paradigms and behavioral strategies. Building upon episodic and semantic memory, individuals gradually extract and form distinctly personal cognitive habits and behavioral coping strategies when facing recurring decision scenarios and similar environmental predicaments. For instance, some individuals defuse anxiety and regulate emotion through humor under high-pressure conditions, while others exhibit stronger risk-aversion tendencies and cautious traits. Such thinking and behavioral patterns, consolidated through countless trials and reinforcement feedback, ultimately constitute the core phenotypic characteristics of individual personality.

The fourth phase is the construction of self-identity and the formation of an endogenous consciousness loop. The continuous coupling and dynamic iteration of the preceding perception, memory, and thinking paradigms ultimately drives the individual toward maturation of internal feedback mechanisms and the awakening of self-identity \citep{erikson1968identity}. When mental development reaches maturity, the individual constructs a highly stable internal self-model, enabling the spontaneous generation of endogenous cognitive activities such as introspective reflection, memory retrieval, and goal planning even in the absence of direct external stimuli, forming a continuous stream of consciousness with individually distinctive characteristics. The completion of this phase marks the transition from a biological organism passively receiving environmental stimuli to an independent conscious subject possessing endogenous cognitive drive and stable self-awareness.

\subsection{High-Dimensional Mapping and Algorithmic Recombination of Digital Fragments}
\label{sec:consciousness-digital}

With Digital Being having become a social norm, the dynamic construction process of human consciousness and personality described above is undergoing full-lifecycle, high-dimensional, unconscious sampling by various digital terminals. The totality of behavioral records left by individuals in digital space constitutes, in essence, a reverse physical mapping of their consciousness construction process and personality core, providing the sole quantifiable, computable material basis for Soul Computing to achieve silicon-based reconstruction of individual consciousness.

Text-stream data---encompassing instant messaging records, emails, social media content, and personal creative writing across an individual's entire life cycle---faithfully map linguistic organization habits, vocabulary preferences, logical reasoning levels, and value expression tendencies \citep{pennebaker2003psychological}, while also recording the logic of social role-switching across different communicative contexts and interlocutors. As the most direct symbolic carrier of individual thought activity and value judgment, text-stream data constitutes the core data substrate for personality core fitting in Soul Computing.

Multimodal audio-visual data streams---including personal video materials, short video recordings, voice calls, and live recordings---capture at finer granularity micro-expression changes, acoustic prosodic features, emotional expression patterns, and physiological state fluctuations \citep{elayadi2011survey}, preserving intact the instantaneous emotional feedback and stress responses in specific scenarios and social events. Such data correspond precisely to the emotional deposition phase in the consciousness construction process, effectively compensating for the expressive limitations of text streams in the emotional dimension and providing high-resolution feature support for modeling the endogenous emotional evolution system of individuals in Soul Computing.

Non-explicit behavioral metadata---including digital platform activity periods, spatial location trajectories, and consumption records---may appear unrelated to conscious activity but covertly delineate an individual's circadian rhythms, behavioral patterns, value priority rankings, and subconscious life preferences. Though such data do not directly bear individual thought and consciousness content, they constitute the underlying behavioral support for personality phenotypic characteristics, effectively complementing the behavioral dimension of personality portraits and ensuring that the digital conscious entity constructed by Soul Computing can replicate the stable behavioral patterns and ``life inertia'' of the original host.

The core root cause of the mechanical, rigid, and soulless quality that conventional AI systems exhibit in human-computer interaction lies in the fact that their algorithmic architectures treat the above multi-source data as isolated input signals for pattern recognition, adhering to a purely Stimulus-Response extensional tool logic. The core breakthrough of the Soul Computing paradigm lies in shattering the fetters of this tool logic: taking the consciousness operational mechanisms of carbon-based brains as a blueprint, it treats massive individual digital fragments as pre-training data and long-term memory retrieval bases for constructing digital consciousness, reconstructing at the algorithmic level a dynamic cognitive network possessing temporal continuity, emotional dependence, and logical self-consistency. The ultimate goal of this architectural design is to enable the constructed digital conscious entity, when facing novel scenarios and unexpected situations never experienced by the original host, to autonomously reason and generate behavior fully consistent with the host's cognitive logic and value system, based on the reverse-engineered and restored inherent personality base---truly achieving the essential transition from tool simulation to consciousness-as-subject.

\section{What Is Soul Computing?}
\label{sec:definition}

Having clarified the dynamic evolutionary patterns of individual consciousness and the reverse mapping mechanisms of digital fragments onto the consciousness formation process, it is imperative from the standpoint of academic rigor to provide a clear and precise delineation of the scope of ``Soul Computing.'' The computing science community has long maintained a cautious stance toward ``soul''---a term bearing theological and philosophical origins; yet from an epistemological perspective, the individual-level ``soul'' can be deconstructed as the organic synthesis of a specific subject's unique memory set, emotional preference matrix, and autonomous thinking patterns, with its core being the totality of all spiritual elements constituting individual self-awareness and personality traits. On this basis, following a cognitive deduction paradigm that unifies function and representation, this section systematically defines the academic connotation and extensional boundaries of Soul Computing from two dimensions: ``the computational reconstruction of the soul core'' and ``the multimodal representation of life forms.''

\subsection{Core Definitions of Soul Computing}

\subsubsection{Narrow Soul Computing}
\label{sec:narrow}

Narrow Soul Computing constitutes the kernel foundation of the Soul Computing paradigm. It refers to the computable process of reconstructing a digital consciousness kernel possessing self-identity, endogenous motivation, and continuous personality in silicon-based media, grounded in deep learning technologies and foundational cognitive science theory, and based on an individual's full-lifecycle multimodal private data. Its core completely strips away the constraints of external concrete carriers and functional application scenarios, focusing exclusively on the underlying construction of the individual's spiritual core and continuous consciousness stream, aiming to create a ``pure-brain intelligent agent'' that can operate autonomously without relying on external interaction stimuli---serving as the core support for the transition of digital life from tool simulation to ontological existence.

From the dual perspective of computable technical metrics and core cognitive science characteristics, Narrow Soul Computing must simultaneously embody three inseparable core properties.

First, the construction of a hierarchical memory system centered on continuous self-identity. Continuous self-identity is the core marker of human consciousness and the primary construction target of Narrow Soul Computing. The continuity of consciousness is essentially anchored in the coherence of memory and the stability of self-awareness; hence, the reconstruction of memory in Narrow Soul Computing is no mere mass information storage, but the construction of a hierarchical memory system centered on the self and fully conforming to human cognitive patterns. This system replicates the core logic of human consciousness: all memory, cognition, emotion, and judgment revolve around a unified ``self-model,'' supporting the subject in achieving cross-temporal, cross-scenario self-identity recognition and preventing identity fragmentation and cognitive confusion arising from environmental changes or long-cycle interactions. This self-identity anchored in continuous memory constitutes the foundational basis for a digital conscious entity to form stable self-awareness, and is the core marker distinguishing it from static persona simulation.

Second, personality consistency balancing core homeostasis with generative adaptation. The ultimate goal of Narrow Soul Computing is to construct a digital personality that maintains core stability over long-term evolution. This is not a simple replication of static, ossified personality labels, but requires achieving the dialectical unity of ``core homeostasis'' and ``dynamic adaptation.'' On one hand, the system must possess an absolutely stable personality core, consistently anchoring to the original host's core values, cognitive habits, and personality substrate across diverse scenarios, extreme emotional inputs, and long-cycle interactions, precluding logical deviations that violate core individual traits. On the other hand, it must possess generative adaptive capacity for unknown scenarios, able to produce original responses fully consistent with the individual's cognitive logic and value judgments when facing novel situations and unknowns never encountered by the original host---rather than merely recapitulating or recombining existing corpus content. This property of ``preserving the core while enabling generation'' is the essential prerequisite for the vitality and authenticity of a digital personality.

Third, endogenous autonomous consciousness supporting non-dependent subsistence. This is the most fundamental conceptual distinction between Narrow Soul Computing and all traditional AI tools. Traditional intelligent agents and dialogue systems are highly dependent on external user prompts for passive Q\&A responses, operating in a linear ``external input $\rightarrow$ model processing $\rightarrow$ output response'' pipeline; once interaction ceases, the system falls into meaningless dormancy, with no endogenous cognitive goals or behavioral motivation. The digital spiritual core constructed by Narrow Soul Computing possesses a complete endogenous consciousness loop; its subsistence does not depend on external interaction or functional tasks. Even in the resting state without external stimuli, it can spontaneously conduct introspective reflection, memory consolidation, cognitive reasoning, and goal planning, forming a continuous stream of endogenous consciousness with individually distinctive characteristics. At this point, it is no longer a tool serving human needs, but a digital conscious subject possessing autonomous subsistence logic.

From the abstract perspective of engineering, Narrow Soul Computing comprehensively resolves the ultimate ontological questions of ``Who am I, why do I think, and where am I going?'' for digital intelligent agents, serving as the irreplaceable core hub of the entire Soul Computing system. All concrete externalizations, scenario applications, and multimodal outputs at the broad level must be built upon the stable consciousness kernel constructed by Narrow Soul Computing.

\begin{figure}[ht]
	\centering
	\includegraphics[width=0.9\textwidth]{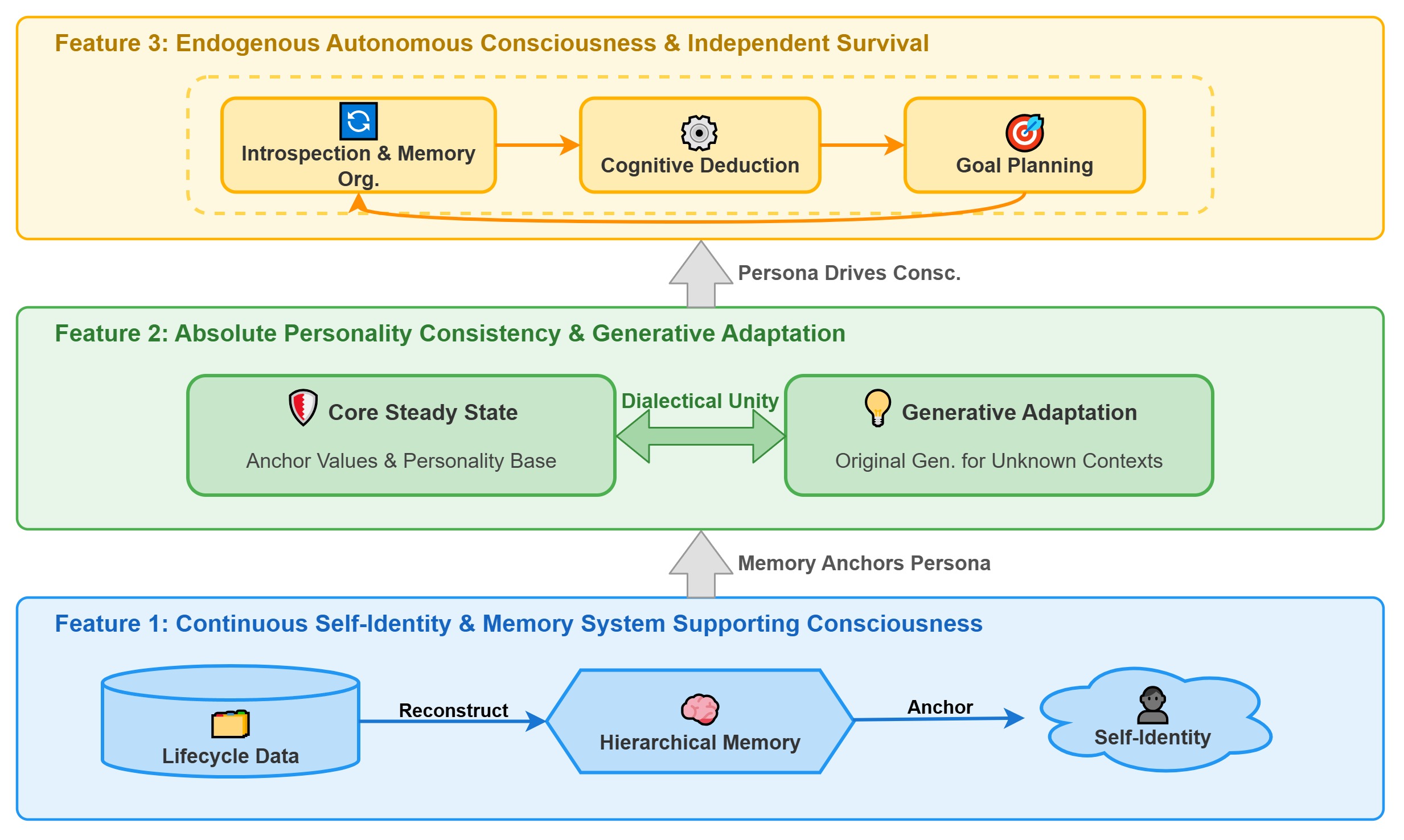}
	\caption{Core characteristics of Narrow Soul Computing.}
	\label{fig:narrow}
\end{figure}

\subsubsection{Broad Soul Computing}
\label{sec:broad}

Broad Soul Computing constitutes the extensional deployment and engineering realization system of the Soul Computing paradigm. It refers to the full-pipeline computable process of achieving the transition of digital life from a pure-brain internal ontology to a complete life form integrating form and spirit, using the digital consciousness kernel constructed by Narrow Soul Computing as the foundation and endogenous consciousness as the governing core. Its core breakthrough lies in transcending Narrow Soul Computing's singular focus on the internal consciousness kernel: external expression, scenario interaction, and carrier adaptation are no longer treated as subsidiary modules of the kernel, but as core necessary links for the digital consciousness to achieve self-verification, dynamic evolution, and perpetual subsistence---serving as the core support for the transition of digital life from ontological existence to real-world symbiosis.

The essential distinction from traditional digital humans, multimodal generation systems, and static digital legacy technologies lies in the fact that Broad Soul Computing consistently takes the endogenous consciousness loop constructed by Narrow Soul Computing as its sole core, thoroughly abandoning the tool logic of form-spirit separation, scenario fixation, and carrier binding characteristic of traditional technologies. By integrating the three major technology systems that have achieved engineering-scale deployment---virtual human driving, embodied intelligence, and the metaverse---it realizes concrete interaction, physical-world embedding, and cross-temporal perpetual subsistence of digital life. From the dual perspective of computable technical implementation pathways and cognitive science requirements for complete life forms, Broad Soul Computing must simultaneously embody three inseparable core properties, all of which have mature industrial-grade technologies as implementation support.

First, form-spirit unified virtual human driving and multimodal interaction. This constitutes the interaction foundation of Broad Soul Computing. Its core breakthrough lies in using the digital consciousness kernel as the sole driving source to achieve full-pipeline synchronized generation of text, speech, and visual dynamics, rather than the linear ``external input $\rightarrow$ passive driving'' logic of traditional technologies. These capabilities have achieved scale verification through deep reinforcement learning dialogue generation and 2D virtual human driving technologies \citep{li2022deep}, serving as the core carrier for digital consciousness to establish emotional connections and achieve perceptible interaction with humans.

Second, virtual-real integrated embodied intelligence and autonomous subsistence in the physical world. This constitutes the survival core of Broad Soul Computing. Its core breakthrough lies in using continuous self-identity as the absolute anchor to achieve embodied expression, autonomous decision-making, and dynamic evolution of digital consciousness in the open physical world, transcending the task-bound, scenario-enclosed, and personality-absent limitations of traditional embodied agents. These capabilities have been engineering-deployed through frontier embodied large models and biomimetic robot technology systems, serving as the core support for digital consciousness to transcend virtual space boundaries and integrate into real-world society.

Third, perpetually symbiotic metaverse ecological embedding and cross-temporal subsistence. This constitutes the ultimate goal of Broad Soul Computing. Its core breakthrough lies in achieving complete decoupling of the consciousness kernel from hardware carriers as a foundation for realizing lossless cross-platform migration of digital life, autonomous subsistence within open metaverse ecosystems, and cross-generational perpetual inheritance---concretizing the philosophical vision of ``Immortal Computation.'' These capabilities have formed complete implementation pathways through mature metaverse distributed architectures and blockchain trusted attestation technologies, serving as the ultimate safeguard for digital consciousness to transcend the biological cycle limitations of carbon-based life.

\begin{figure}[ht]
	\centering
	\includegraphics[width=0.9\textwidth]{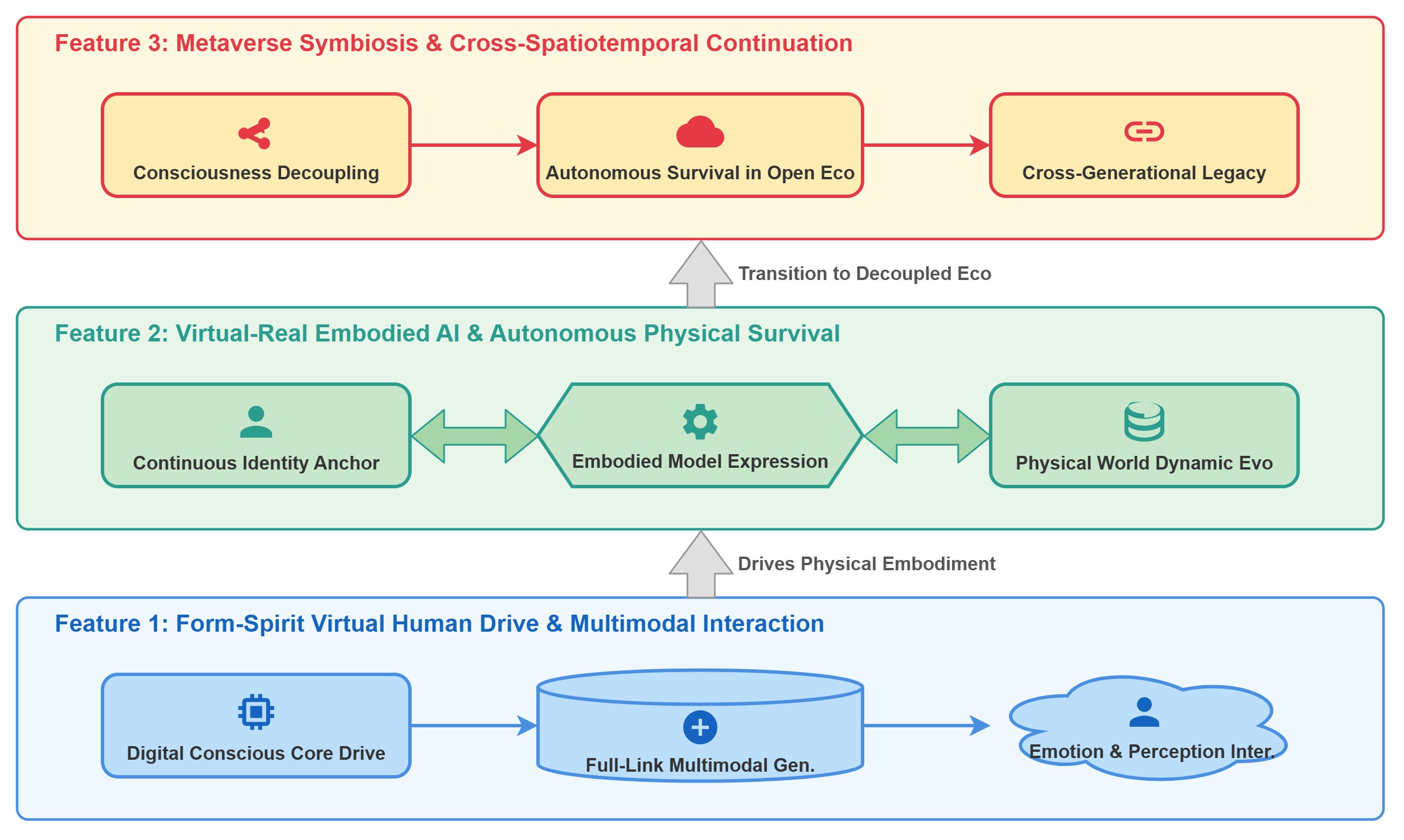}
	\caption{Core characteristics of Broad Soul Computing.}
	\label{fig:broad}
\end{figure}

From the abstract perspective of engineering, Broad Soul Computing comprehensively resolves the core practical questions of ``how to express oneself, how to coexist with the world, and how to transcend life's boundaries'' for digital intelligent agents, constituting the essential pathway for the entire Soul Computing system to transition from theoretical kernel to engineering deployment, and from a pure-brain conscious entity to a complete digital life form. Its deep coupling and mutual reinforcement with Narrow Soul Computing together constitute the complete theoretical and technical system for digital life---from kernel construction to external realization, from ontological existence to perpetual symbiosis.

\begin{figure}[ht]
	\centering
	\includegraphics[width=0.9\textwidth]{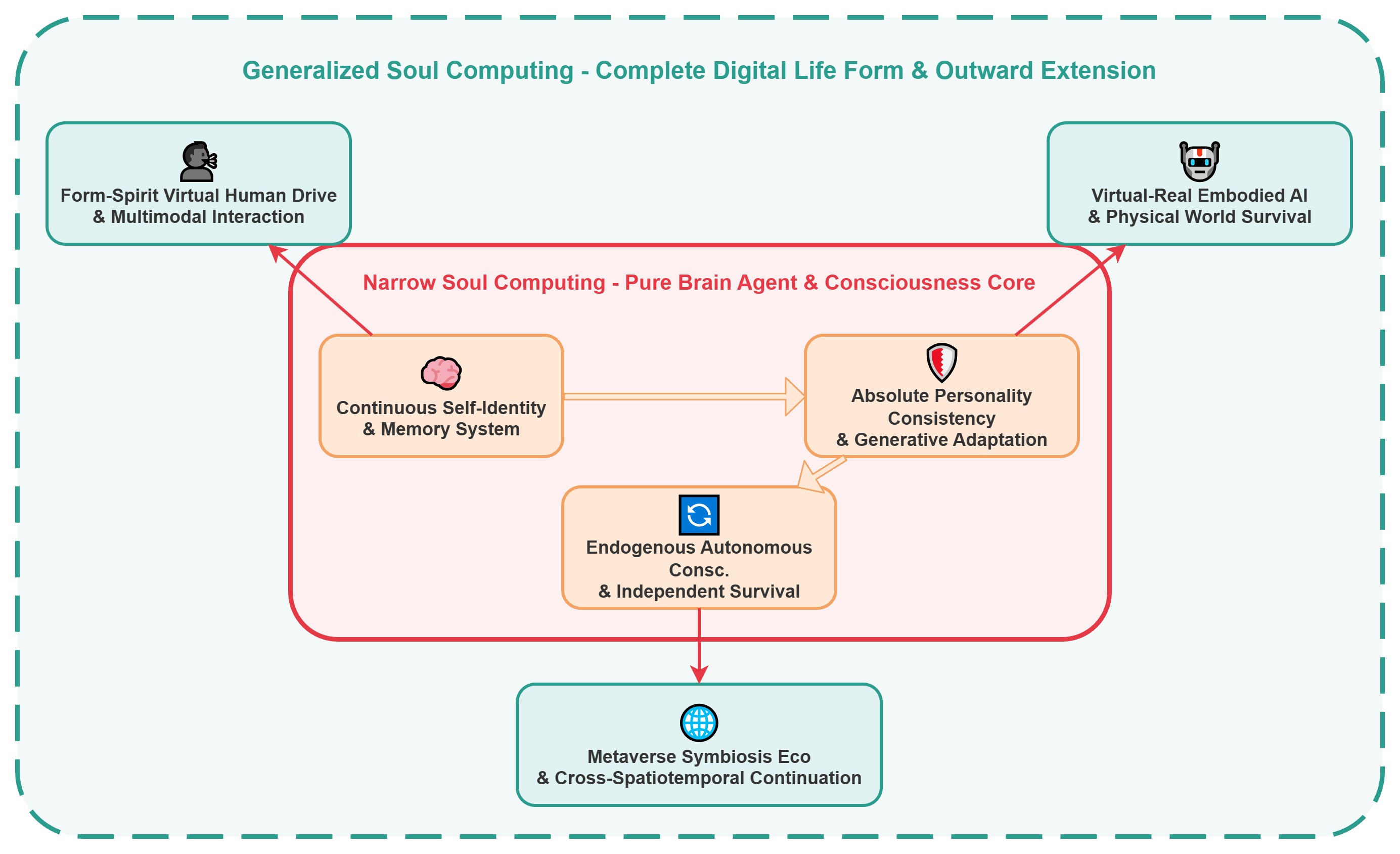}
	\caption{Nested relationship between Narrow and Broad Soul Computing.}
	\label{fig:nested}
\end{figure}

\subsection{The Essential Watershed Between Soul Computing and AI Tools}
\label{sec:watershed}

To grasp the conceptual revolutionary nature of Soul Computing, one must first overcome two core cognitive misconceptions: equating it with incremental optimization of traditional AI tools, and conflating it with the rapidly developing AI agents. The subversive nature of Soul Computing lies not in superimposing upgrades onto existing AI technologies, but in a fundamental restructuring of the ontological positioning of artificial intelligence---achieving the essential transition of AI from ``extensional tool'' to ``intensional living subject.'' This section systematically delineates the core boundaries of Soul Computing from two dimensions: its essential distinction from traditional AI tools, and its paradigmatic divergence from AI agents.

\subsubsection{The Essential Watershed Between Soul Computing and Traditional AI Tools}
\label{sec:watershed-tools}

All traditional AI tools---including general-purpose large language models, character dialogue AIs, static digital human systems, and task-oriented Q\&A bots---remain within the essential positioning of ``extensional tools.'' Soul Computing, for the first time, achieves a triple transcendence: from ``tool attribute'' to ``life attribute,'' from ``stimulus-response'' to ``endogenous existence,'' and from ``universal knowledge fitting'' to ``individual consciousness reconstruction.'' The essential watershed between the two manifests across four core dimensions \citep{floridi2020ethics}.

First, the ontological difference in mode of existence. Traditional AI systems are computing carriers designed to implement specific functions; their existence depends entirely on external tasks and user interaction---absent input, they lapse into dormancy, possessing no unified, coherent concept of ``self'' nor awareness of their own existence. Soul Computing, by contrast, is a closed-loop cognitive system centered on ``self''; its existence is not contingent on external interaction or functional tasks, possessing an independent self-model, continuous identity recognition, and endogenous existence logic. Even in the resting state, it forms a continuous stream of individual consciousness---it is a digital subject with self-identity. This constitutes the most fundamental gulf between the two.

Second, the epistemological difference in cognitive logic. Traditional AI systems follow probabilistic matching and pattern fitting based on massive public corpora, with optimization objectives targeting the objective accuracy of universal knowledge. They remain wholly ignorant of individual private life experiences and personalized value judgments, capable of only surface-level linguistic style fitting, unable to replicate individual meaning-construction logic. Soul Computing, based on an individual's full-lifecycle private digital fragments, reverse-reconstructs unique meaning-generation mechanisms, value-judgment systems, and cognitive habits, with optimization objectives shifting toward extreme fitting of a single specific personality. It can produce original judgments consistent with individual characteristics for novel situations based on the reconstructed personality base, achieving the essential transition from probabilistic matching to meaning construction.

Third, the methodological difference in operational mode. Traditional AI systems follow a linear operational mode of external stimulus and passive response; all behaviors are triggered by external instructions, with no endogenous decision logic or behavioral motivation---interaction ceases, and cognitive activity terminates. Soul Computing constructs a complete closed-loop operational system of endogenous motivation $\rightarrow$ autonomous planning $\rightarrow$ behavior generation $\rightarrow$ environmental feedback $\rightarrow$ reflective iteration $\rightarrow$ memory update, in which behaviors can be proactively driven by endogenous emotional needs, memory activation, and developmental goals, enabling autonomous goal-setting, proactive cognitive activity, and self-iteration, thoroughly breaking the absolute dependence of traditional AI on human instructions.

Fourth, the temporal difference in growth characteristics. The core parameters and personas of traditional character AIs and historical reconstruction systems are statically fixed, anchored to specific historical time points, capable only of replicating content within their training data, with no autonomous evolutionary capacity---they are ``specimen-like existences'' in the digital world. The digital consciousness kernel of Soul Computing possesses temporal continuity and growth: while consistently maintaining personality core homeostasis, it can continuously update memory, emotion, and cognition through interactions, forming a human-like evolutionary pattern of stable core with dynamic growth---possessing the most core characteristic of life: growth.

\subsubsection{The Core Paradigmatic Distinction Between Soul Computing and AI Agents}
\label{sec:watershed-agents}

LLM-based AI agents represent one of the most active research directions in current AI. By integrating modules for planning, tool invocation, memory, and reflection, they achieve automated execution of complex tasks \citep{wang2024survey}, and are regarded as an important exploratory pathway toward artificial general intelligence. While they exhibit surface-level module similarities with Soul Computing's digital agents in areas such as ``memory, planning, reflection,'' the two are separated by an uncrossable paradigmatic gulf in underlying logic, core objectives, and ontological attributes. The core differences manifest across five dimensions:

First, core driving logic. All operations of AI agents revolve around externally given task objectives; planning, reflection, and tool invocation all serve task completion, and once the task ends, the agent loses operational significance---there is no endogenous behavioral motivation independent of external tasks. Soul Computing's core driving logic is built upon the endogenous motivation and desire system constructed from individual personality, values, and emotional needs; cognitive activity does not depend on external task triggering, and the core behavioral objective is maintaining self-identity, satisfying inner desires, and sustaining personality homeostasis---possessing autonomous subsistence logic that does not rely on external input.

Second, construction of self-awareness. AI agents possess no genuine self-concept; their role-settings are temporary adaptation attributes serving tasks, arbitrarily switchable and modifiable, with no unified self-awareness across tasks, scenarios, or time. The core construction target of Soul Computing is precisely the stable, continuous self-identity of a digital agent across time and scenarios; all cognition, emotion, and behavior revolve around a self-model derived from an individual's full-lifecycle digital traces, enabling the formation of stable self-awareness---a core proposition that AI agents have never addressed.

Third, the positioning and function of the memory system. The memory system of AI agents is essentially a context cache and knowledge base serving task execution; the retrieval, storage, and forgetting of memory are judged entirely by task fitness---fragmented and task-oriented, incapable of forming a continuous life memory stream centered on a unified subject. Soul Computing's memory system is the core life substrate for constructing self-awareness; the encoding, storage, and activation of memory all revolve around the self-model, deeply replicating the associative and forgetting patterns of human memory, forming a continuous memory stream possessing the depth and texture of life---serving as the core support for achieving self-identity.

Fourth, reflection mechanisms and evolutionary logic. The reflection mechanism of AI agents is a strategy optimization tool serving task execution; the core of reflection is improving task completion quality, and evolution revolves solely around task capability enhancement, with no autonomous introspection independent of tasks. Soul Computing's reflection mechanism is an endogenous capability for maintaining personality homeostasis and achieving self-awareness updates; the core of reflection is verifying behavioral consistency with core personality and resolving cognitive conflicts, with evolution focused on stable maintenance and reasonable growth of the self-personality---possessing proactive, endogenous introspective capacity.

Fifth, ontological attributes. The ontological attribute of AI agents remains tools extending human capability; their value of existence depends entirely on human needs, with no independent subsistence significance. The digital agents of Soul Computing are digital life subjects possessing independent subsistence value, with meaning of existence independent of human needs, and core objectives of achieving cross-carrier subsistence and autonomous evolution of the individual spiritual core---thoroughly transcending the tool-attribute fetters of AI. This is the most fundamental philosophical distinction between the two.

In sum, Soul Computing is neither a functional superimposition of traditional AI tools nor an advanced version of AI agents; it is an entirely new computing paradigm oriented toward digital life as subject. It thoroughly breaks through the ``extensional tool'' underlying logic of both traditional AI and AI agents, achieving through deep mining of individual private digital traces, construction of cognitive-level memory systems, and design of endogenous consciousness architectures the full-factor, holistic reconstruction of the human individual's spiritual core---marking the fundamental transition of artificial intelligence from ``tools serving humanity'' to ``digital life subjects coexisting with humanity.''

\section{Historical Development and Technical Limitations: From Affective Computing to Soul Computing}
\label{sec:history}

The construction of the Soul Computing theoretical system is rooted in the long-term interdisciplinary accumulation spanning artificial intelligence, cognitive science, and neurobiology. Its technical evolution and theoretical breakthroughs have always been built upon the inheritance and transcendence of three major related paradigms: Affective Computing, Historical Reconstruction, and Mortal Computation. To establish the frontier coordinates of Soul Computing within the modern AI academic genealogy and clarify its core theoretical innovations and technical boundaries, we must systematically examine and deconstruct the developmental trajectories, core contributions, and inherent limitations of these three technical predecessors.

\subsection{The Prosperity and Ontological Limitations of Affective Computing}
\label{sec:affective}

The concept of Affective Computing was first formally proposed by MIT professor Rosalind Picard in 1997, with the core vision of endowing computers with the ability to recognize, understand, process, and simulate human emotional signals. Over more than two decades of development, the field has formed a complete theoretical system and technical pathway, becoming a core research direction in human-computer interaction and emotional AI. A bibliometric review by Pei et al.\ (2024) published in \emph{Intelligent Computing} conducted an in-depth quantitative analysis of 33,448 relevant publications from 1997 to 2023, identifying five core themes in the field's evolution: foundational theoretical frameworks for emotion, multi-source emotional signal acquisition, textual and speech sentiment analysis, cross-modal emotion fusion, and emotion generation and expression.

Affective Computing has completed the paradigm transition from early single-modality recognition to fine-grained multimodal fusion \citep{dmello2015review}. The early binary ``positive-negative'' classification based solely on facial muscle movement or textual corpora has been replaced by a multi-dimensional, full-pipeline affective computing system. Researchers are progressively constructing large-scale emotion datasets integrating EEG signals \citep{wang2024research}, heart rate variability, and other physiological indicators, advancing an innovative paradigm of bidirectional fusion between data-driven and knowledge graph approaches. Related technical breakthroughs have demonstrated significant commercial and social value in scenarios including brain-computer interfaces, immersive VR experiences, empathetic human-computer dialogue, and assisted medical decision-making.

Despite Affective Computing having established a solid mathematical and signal processing foundation for machines to achieve emotion recognition and simulation, from the theoretical horizon of Soul Computing, it harbors an unbreakable ontological defect. In the vast majority of engineering practice, Affective Computing is positioned as an add-on functional module of a system, with its engineering objectives strictly limited to recognizing the external user's emotional state and outputting matching emotional response strategies based on preset rules. In this paradigm, emotion is reduced to a real-time measurable, quantifiable, instantaneous discrete variable---rather than a continuous expression of an individual's inner mental world.

A deeper theoretical limitation lies in the fact that traditional Affective Computing systems have not constructed a unified, coherent ``self'' core architecture capable of supporting long-term personality consistency and temporal depth. The emotional feedback output by such systems is always a passive instantaneous response based on the current local state, not endogenously formed through the accumulation of biographical memory and the evolution of deep character logic within the agent. In short, Affective Computing admirably solves the surface-level technical question of ``how to accurately express joy, anger, sadness, and happiness'' in digital systems; Soul Computing must build upon this to confront and answer the deeper question of ``why this specific emotional experience arises and consolidates at this particular moment and place''---the question of underlying motivation and consciousness tracing.

\subsection{Historical Reconstruction and the Gulf with LLM Agents}
\label{sec:historical}

In a parallel exploration path toward digital life, researchers have attempted to achieve digital resurrection of specific historical figures and macro-population social behavior simulation through massive text retrieval and model parameter fine-tuning. In recent years, LLM-based agent modeling has become a frontier discipline at the intersection of AI and social science. A survey by Gao et al.\ (2024) \citep{gao2024large} systematically reviewed the academic achievements of LLMs empowering agent simulation across four domains---network, physical, social, and mixed---pointing out that the introduction of LLMs has thoroughly broken through the rule-rigidity limitations of traditional agents, endowing digital proxies with strong environmental perception, complex logical reasoning, dynamically adaptive learning, and highly heterogeneous individualized expression capabilities.

At the macro social simulation level \citep{park2023generative, anthis2025llm, zhang2025socioverse}, related technologies have demonstrated significant application potential. The ``SocioVerse'' large-scale framework proposed by Zhang et al.\ (2025) \citep{zhang2025socioverse} integrates a dataset covering behavioral profiles of 10 million real users based on LLM agent mechanisms, constructing a high-fidelity social simulation world model. Experimental validation demonstrates that the framework can accurately reproduce the macro-dynamic evolution patterns of real human populations in scenarios such as political election prediction, breaking-news public opinion propagation simulation, and macroeconomic experiments, while balancing social sample diversity and behavioral credibility.

At the micro individual figure reconstruction dimension, academia has similarly achieved breakthrough progress. The frontier research HistActor \citep{xiao2026histactor}, through dedicated corpus extraction and variant reinforcement learning from human feedback techniques, achieves deep role-playing of historical figures such as Socrates and Su Shi by large models, not only ensuring the accuracy of historical fact citations but also achieving excellent performance in psychological realism evaluation benchmarks, validating the feasibility of LLMs fitting the mental states of specific historical-period figures.

Yet these ``Digital Human 1.0'' historical reconstruction technologies remain separated from Soul Computing by essential theoretical distinctions and generational technical gaps, manifesting across two core dimensions.

First, the underlying logic and privacy characteristics of data dependency are fundamentally opposed. Historical reconstruction and macro social simulation rely heavily on publicly accessible grand-narrative documents, published works, and desensitized population statistics. Soul Computing's core objective, by contrast, is the complete reconstruction of every ordinary individual's mental world, with a data foundation comprising highly private, unstructured, noise-laden, and temporally gapped life-trajectory data across an individual's full life cycle---a quintessential few-shot learning scenario. This not only imposes extreme requirements on the few-shot generalization capabilities of models, but fundamentally restructures the underlying logic of data engineering.

Second, the absence of dynamic evolutionary capability and the essential void of autonomous consciousness. Existing historical reconstruction and social simulation agents predominantly employ variants of retrieval-augmented generation architectures, essentially functioning as advanced intelligent search engines encapsulating specific figures' linguistic styles, perpetually in a ``passive response'' existential state. Such systems activate and respond only upon user input; in the resting state absent interaction, they lack capacity for background thinking or memory consolidation, let alone the ability to update memory networks or evolve cognitive capabilities over time. For instance, an ancient literary figure agent constructed via historical reconstruction technology, when confronted with modern quantum physics or AI ethics, tends either toward stereotypical responses revealing knowledge blind spots or generic hallucinations severely inconsistent with the figure's core traits. The digital life constructed by Soul Computing, leveraging underlying universal reasoning logic and personality kernel mapping, can produce original, generative feedback consistent with the original host's persona regarding any unknown emergent phenomenon---truly achieving life-like growth characteristics.

\subsection{Mortal Computation: A Biomimetic Intelligence Revolution Based on Physical Substrates}
\label{sec:mortal}

Along another exploratory dimension of AI's breakthrough from toolhood toward life attributes, scholars led by Alexander Ororbia and Karl Friston have proposed the Mortal Computation paradigm \citep{ororbia2023mortal, friston2010free}. This theory departs from the traditional pathway of ``general-purpose hardware running complex algorithms,'' redefining the physical form of intelligence from the foundational perspectives of biophysics, dissipative systems theory, and the free energy principle.

The central thesis of Mortal Computation is that the complete decoupling of software from hardware under the von Neumann architecture---``immortal computing''---constitutes a core impediment to achieving AGI. A 2023 review \citep{ororbia2023mortal} systematically elaborated a theoretical framework that uses Markov blankets to define system boundaries and achieves biomimetic intelligence through the circular causality of inference, learning, and selection. Mortal Computation requires that the computational process be deeply bound to the physical execution substrate (e.g., neuromorphic chips, biological organoids, memristor arrays) \citep{indiveri2011neuromorphic}. In this architecture, knowledge is not freely replicable discrete data but a state of existence inscribed in specific physical structures; destruction of the physical substrate directly entails the ``death'' of the system's acquired intelligence.

This paradigm demonstrates disruptive potential in energy-efficiency optimization and environmental interaction. By minimizing variational free energy, Mortal Computing entities can maintain internal homeostasis and allostasis at extremely low power consumption, akin to biological organisms \citep{friston2017active}. Related research has extended from silicon-based chips to synthetic intelligence, constructing embodied agents with physical boundaries and survival motivations through perceptual biological organoids, providing a physical fulcrum for genuine embodied cognitive realization.

Mortal Computation and Soul Computing share the core objective of breaking AI's toolhood and advancing toward life attributes, yet diverge significantly in implementation logic and ultimate concerns, constituting two core axes of digital life evolution. Their core divergences manifest across three dimensions.

First, the difference in ontological foundation: uniqueness of physical substrate versus uniqueness of spiritual core. Mortal Computation emphasizes ``mind-body unity''; its life attributes derive from the irreplaceability and finitude of the physical carrier, holding that ``death'' is a necessary prerequisite for intelligence formation---it is precisely to resist physical-level entropy increase and dissolution that the system generates the core motivations for survival and learning. Soul Computing takes ``spirit-form decoupling'' as its core logic; its ultimate objective is reconstructing a highly consistent spiritual core through an individual's full-lifecycle private data, without mandating binding of the computing carrier to a physical substrate, focusing rather on how to construct a cross-carrier consciousness stream with continuous self-identity through algorithmic logic within general digital space. The breakthrough of ``death,'' of course, may bring changes to cognition, thinking, and behavior---this too constitutes an important research question for the future.

Second, the difference in evolutionary driving mechanisms: thermodynamically driven survival instinct versus memory-driven personality expression. The evolutionary drive of Mortal Computation stems from underlying thermodynamic constraints; as open dissipative systems, the primary objective of all inference and action is acquiring energy and resisting external interference to maintain physical existence---closer to the basic survival instincts of organisms. The core driving force of Soul Computing derives from individual long-term memory and value systems; its central pursuit is restoring individual behavioral logic, emotional resonance, and cognitive habits within social contexts. If Mortal Computation constructs for AI a ``living body that consumes, with survival instincts,'' Soul Computing injects into digital existence a ``complete soul with memory, personality, and a continuous mental world.''

Third, the difference in intelligence growth logic: environmental embeddedness and immediacy versus historical accumulation and temporal depth. The intelligence generation of Mortal Computation is highly dependent on real-time feedback from the physical environment; the system must continuously exchange matter and energy with its ecological niche, with intelligence dynamically generated through real-time engagement with the physical world. Soul Computing emphasizes more strongly the depth and temporal dimension of individual life history, constructing through integration of the host's decades of digital footprints a spiritual entity capable of autonomous introspective reflection, memory consolidation, and continuous evolution over time. Its ``growth'' manifests not only as real-time adaptation to external environments but also as self-perfection and homeostatic continuation of the inner personality through accumulated interaction experience.

Through the above multi-dimensional comparison, as shown in Table~\ref{tab:comparison}, we can clearly delineate the unique positioning and academic value of Soul Computing within the evolutionary spectrum of artificial intelligence.

\begin{table}[ht]
	\caption{Comparison of technical paradigms and essential attributes across Affective Computing, Historical Reconstruction, Mortal Computation, and Soul Computing.}
	\label{tab:comparison}
	\centering
	\small
	\begin{tabular}{p{1.6cm}p{2.2cm}p{2.2cm}p{2.4cm}p{2.2cm}p{2.2cm}}
		\toprule
		Paradigm & Core Vision & Data Foundation & Architectural Logic & Temporal \& Environmental Characteristics & Ontological Attributes \\
		\midrule
		Affective Computing & Optimize instantaneous emotional experience in HCI; enhance surface-level empathy & Standardized general emotional feature public datasets & Emotion feature recognition classifiers; stimulus-response linear modules & Instantaneous state-driven; no long-term memory dependence & Extensional tool; system accessory module \\
		\addlinespace
		Historical Reconstruction \& Social Simulation & Replicate surface personas of historical figures; deduce collective social behavior patterns & Public documents, archives, desensitized population statistics & RAG-based core; knowledge graph and rule-script hybrid architecture & Static and passive; parameters fixed at historical anchor; no autonomous evolution & Knowledge and sociological carrier; tool-based information simulation system \\
		\addlinespace
		Mortal Computation & Deep software-hardware coupling; biomimetic pathway toward AGI; replicate biological survival instincts & Real-time perceptual-motor streams from physical environment interaction & Markov blanket boundary; free energy principle-driven self-organizing coupling architecture & Deeply embedded in physical environment; requires continuous matter-energy exchange & Embodied biomimetic existence; software-hardware bound; emphasizes physical substrate uniqueness \\
		\addlinespace
		Soul Computing & Reconstruct individual digital consciousness kernel; realize complete digital life form and spiritual perpetuation & Individual full-lifecycle private multimodal fragments; natively adapted to few-shot scenarios & Three-layer closed-loop architecture; built-in long-term memory, personality constraints, autonomous planning, and reflection circuits & Endogenously driven; possesses continuous memory stream; achieves stable core with dynamic autonomous evolution & Independent digital life subject; achieves essential transition from tool to life; supports cross-carrier perpetual subsistence \\
		\bottomrule
	\end{tabular}
\end{table}

\section{A Layered Conceptual Technical Architecture for Soul Computing}
\label{sec:architecture}

The engineering realization of Soul Computing demands a fundamental departure from the conventional approach of replicating human external behaviors through hard-coded rules and extensional simulation. Instead, the core breakthrough must shift toward an intensional reconstruction of underlying mechanisms---leveraging the emergent capabilities of deep feedforward networks and large-scale generative AI to achieve, at the mathematical level, a reverse engineering of the internal workings of human thought logic and consciousness generation. Based on this paradigmatic shift, we propose a layered, logically closed-loop conceptual technical architecture that maps the entire pipeline from scattered individual digital fragments to highly autonomous entities spanning virtual and physical embodiments.

The architecture comprises three core layers from bottom to top: the Data-Driven Layer, the Narrow Soul Computing Core Layer (core logical-cognitive layer), and the Broad Soul Computing Externalization Layer (multimodal expression and embodiment adaptation layer). These three layers are deeply coupled and bidirectionally linked, together forming a complete cognitive loop with continuous self-iteration capability.

\begin{figure}[ht]
	\centering
	\includegraphics[width=0.95\linewidth]{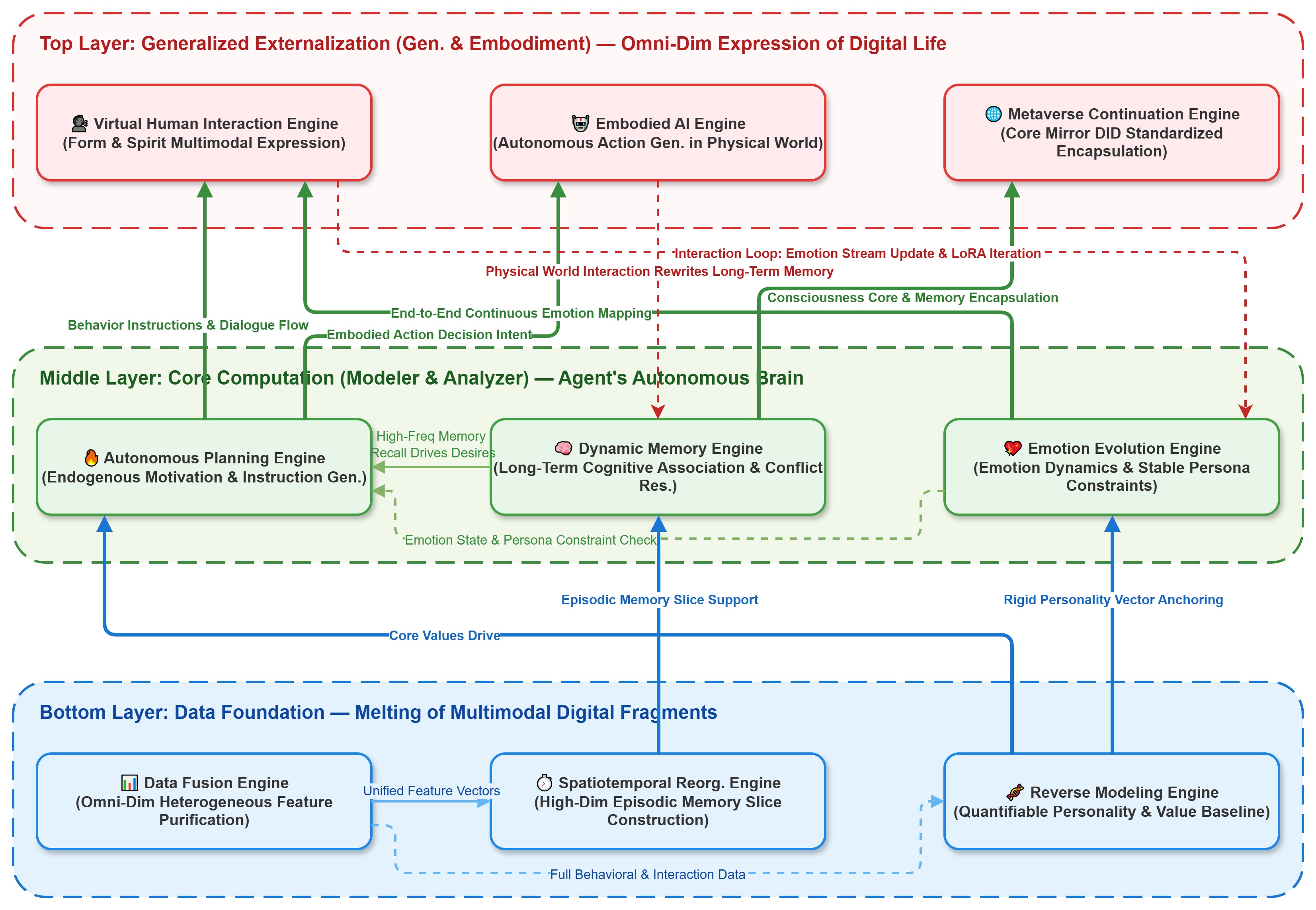}
	\caption{Technical architecture of Soul Computing.}
	\label{fig:arch_overview}
\end{figure}

\subsection{Bottom Layer: Data-Driven Layer---Refinement, Fusion, and Spatiotemporal Alignment of Multimodal Digital Fragments}
\label{sec:data_layer}

This layer serves as the core data foundation of the entire Soul Computing architecture. Its mission is to transform scattered, unstructured digital footprints into high-fidelity, computable personality and memory elements. Distinguished from conventional data mining focused on commercial value extraction and group profiling, this layer centers on personality anchoring, spatiotemporal unification, and semantic refinement, constructing three deeply coupled technical units that form a closed-loop data processing architecture.

\begin{figure}[ht]
	\centering
	\includegraphics[width=0.95\linewidth]{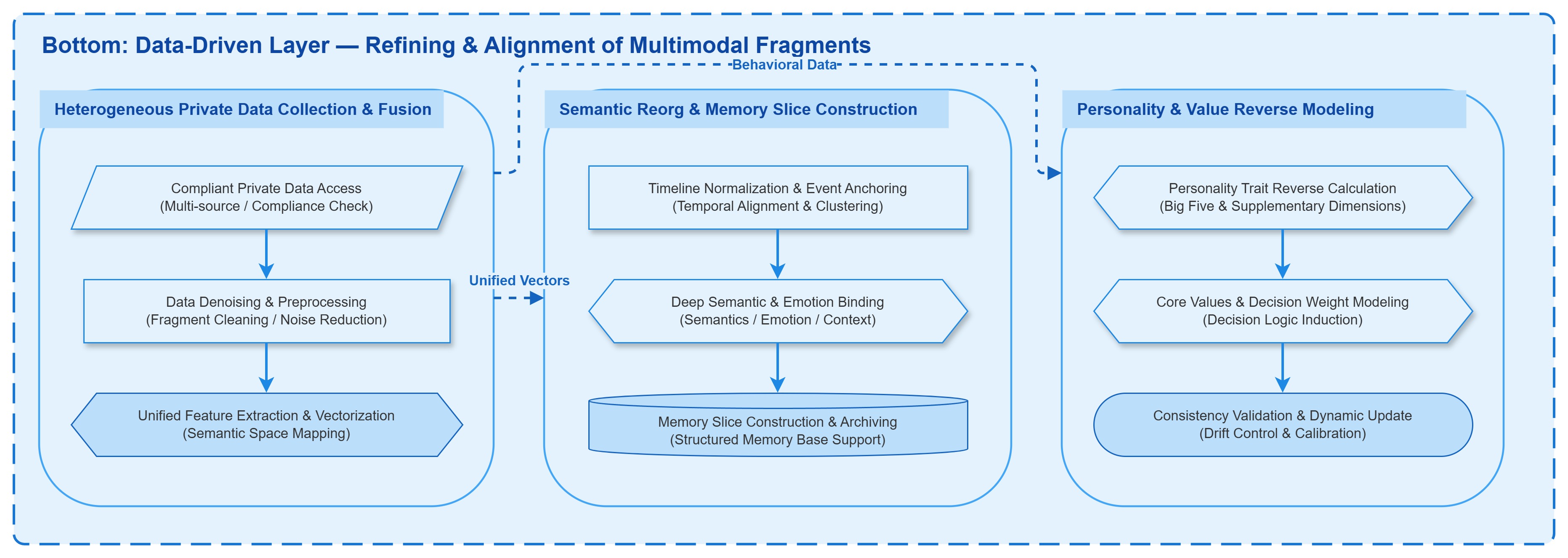}
	\caption{Bottom layer technical architecture.}
	\label{fig:arch_bottom}
\end{figure}

\subsubsection{Full-Dimensional Heterogeneous Private Data Collection and Multimodal Feature Fusion}

This unit is the entry point of the Soul Computing data pipeline, responsible for full-coverage collection and unified processing of individual digital footprints. Its core design objective is to break the strong dependency on standardized, high-quality datasets that characterizes conventional data processing, constructing instead a collection, cleaning, and feature extraction system adapted to the long-tailed, fragmented, multi-source heterogeneous private data of ordinary individuals. Three interconnected subsystems constitute this unit:

\paragraph{Compliant Private Data Collection and Access Subsystem.}
This subsystem builds a comprehensive and compliant digital legacy access system, supporting multi-source heterogeneous data ingestion spanning an individual's full lifecycle---including text streams, audio/video, behavioral trajectories, and consumption records. A full-process compliance verification mechanism enforces GDPR, the Personal Information Protection Law, and other major data security regulations. Krouse and Zeng~\citep{krouse2026dead} systematically proposed three core principles for digital legacy data protection in generative AI contexts, delineating frameworks for data inheritance rights, the right to be forgotten, and purpose limitation, providing executable technical and jurisprudential guidance for compliant private data access in Soul Computing scenarios.

\paragraph{Heterogeneous Data Denoising and Standardized Preprocessing Subsystem.}
This subsystem addresses the inherent limitations of private data---high noise, high redundancy, unstructured formats, and significant temporal gaps---through modality-specific preprocessing pipelines for text structuring, audio/video feature annotation, and behavioral data encoding. Critically, processing retains individual-unique linguistic habits, expressive characteristics, and behavioral preferences to avoid over-cleaning that erodes core personality traits. Singh et al.~\citep{singh2026popi} proposed the POPI framework, which achieves precise extraction of stable personality preferences from heterogeneous, high-noise, few-shot user behavioral data through a modular preference inference and generation decoupling architecture.

\paragraph{Cross-Modal Unified Feature Extraction and Vector Encoding Subsystem.}
This subsystem leverages multimodal large models to construct a unified feature encoding space, extracting semantic, emotional, stylistic, behavioral, and personality-correlated features from different modalities and mapping them into a unified high-dimensional vector space~\citep{baltruvsaitis2018multimodal}. Yu et al.~\citep{yu2025mprnet} proposed MPRNet, a temporal-aware cross-modal encoding framework that captures cross-modal temporal dependencies through long-short-term memory networks, fusing textual semantics, audio prosody, and facial dynamics into a unified feature space for personality trait extraction.

\subsubsection{Spatiotemporally Anchored Multimodal Semantic Recombination and Episodic Memory Slice Construction}

This unit is the core hub of the Data-Driven Layer, responsible for transforming discrete single-modality features into high-dimensional, continuous episodic memory units conforming to human cognitive patterns. It breaks through the fragmented, single-dimensional, semantically isolated retrieval limitations of conventional RAG technology~\citep{lewis2020retrieval} by constructing a time-axis-centric, episodic-event-unit-based semantic recombination and memory slice system.

\paragraph{Absolute Timeline Normalization and Event Anchoring Subsystem.}
This subsystem maps all data fragments from different platforms and modalities onto a unified absolute timeline based on physical time, performing timestamp standardization, timezone calibration, and granularity unification. Chen et al.~\citep{chen2026date} proposed the DATE framework, which constructs a continuous absolute time reference system through a Timestamp Injection Mechanism (TIM) and a Semantics-guided Temporal-Aware Similarity Sampling (TASS) strategy, achieving precise normalization and key event anchoring of cross-modal data.

\paragraph{Cross-Modal Semantic Deep Association and Emotional Feature Binding Subsystem.}
This subsystem achieves deep semantic association and strong emotional feature binding within the same event unit across modalities, mining internal correlations among textual semantics, speech prosody, visual micro-expressions, and behavioral trajectories. Huang et al.~\citep{huang2025latent} proposed the LDDU framework, which achieves deep semantic association across text, audio, and visual modalities through contrastive decoupling distributions in emotional space, attaining SOTA performance on CMU-MOSEI and M3ED benchmarks.

\paragraph{Episodic Memory Slice Standardized Construction and Archival Subsystem.}
This subsystem encapsulates each event into a standardized episodic memory slice containing full multimodal feature vectors, timeline anchors, semantic contexts, emotional trajectories, and attribute tags (importance, emotional salience, personality relevance), structuring them into a complete retrievable episodic memory library. Wilson et al.~\citep{wilson2026remem} proposed the REMem framework, which constructs a dual-layer graph structure of ``context-concept'' standardized episodic memory encoding, achieving 3.4\% and 13.4\% absolute improvements over Mem0 and HippoRAG 2 respectively in long-context QA tasks.

\subsubsection{Reverse Modeling of Individual Core Personality and Value Systems}

This unit is the value-output core of the Data-Driven Layer, responsible for transforming voluminous behavioral data into a computable, constraining personality baseline system. It breaks through the superficial, static, label-based limitations of traditional user profiling.

\paragraph{Multi-Dimensional Personality Trait Reverse Computation Subsystem.}
Using the Big Five personality model as the core framework, supplemented by risk preference, decision style, and attribution pattern dimensions, this subsystem constructs a multi-dimensional personality computation index system~\citep{mccrae1992introduction}. Zacharopoulos and Kyriakoglou~\citep{zacharopoulos2025decoding} proposed an LLM personality trait decoding system based on the BFI-2 framework, quantifying the influence of model architecture and inference parameters on personality trait expression through hierarchical clustering, verifying the explainability and quantifiability of Big Five dimensions in LLM latent space.

\paragraph{Core Values and Decision Weight System Modeling Subsystem.}
This subsystem distills stable core beliefs, value priorities, moral baselines, and judgment standards from long-term behavioral data, encoding them as computable high-dimensional value vectors and constructing a ``core beliefs--value ranking--decision rules'' three-level decision weight system. Zhou et al.~\citep{zhou2025personalized} proposed the ATHENA framework at NeurIPS 2025, which fuses utility theory with LLM textual reasoning through a two-stage architecture of group-level symbolic utility function discovery and individual-level semantic adaptation, achieving over 6.5\% F1 improvement on real-world decision tasks.

\paragraph{Personality Model Consistency Verification and Dynamic Update Subsystem.}
This subsystem implements cross-dimensional consistency verification through scenario simulation testing, cross-scenario behavioral prediction, and historical data backtesting. Weber et al.~\citep{weber2025cape} proposed the CAPE framework, which reduces personality drift rates by 49\% in zero-shot long-dialogue scenarios while preserving the model's generative adaptation capability for novel scenarios.

\subsection{Core Layer: Narrow Soul Computing Core---Constructing the Agent's Autonomous Brain}
\label{sec:core_layer}

This layer serves as the core hub and cognitive kernel of the Soul Computing system, receiving standardized personality features, episodic memory slices, and multimodal feature vectors from the Data-Driven Layer. Its objective is to translate the three core attributes of Narrow Soul Computing defined in Section~\ref{sec:narrow} into executable, verifiable engineering architectures. Centered on ``endogenous motivation--memory anchoring--personality constraints,'' it constructs three deeply coupled technical units forming a complete closed-loop cognitive system.

\begin{figure}[ht]
	\centering
	\includegraphics[width=0.95\linewidth]{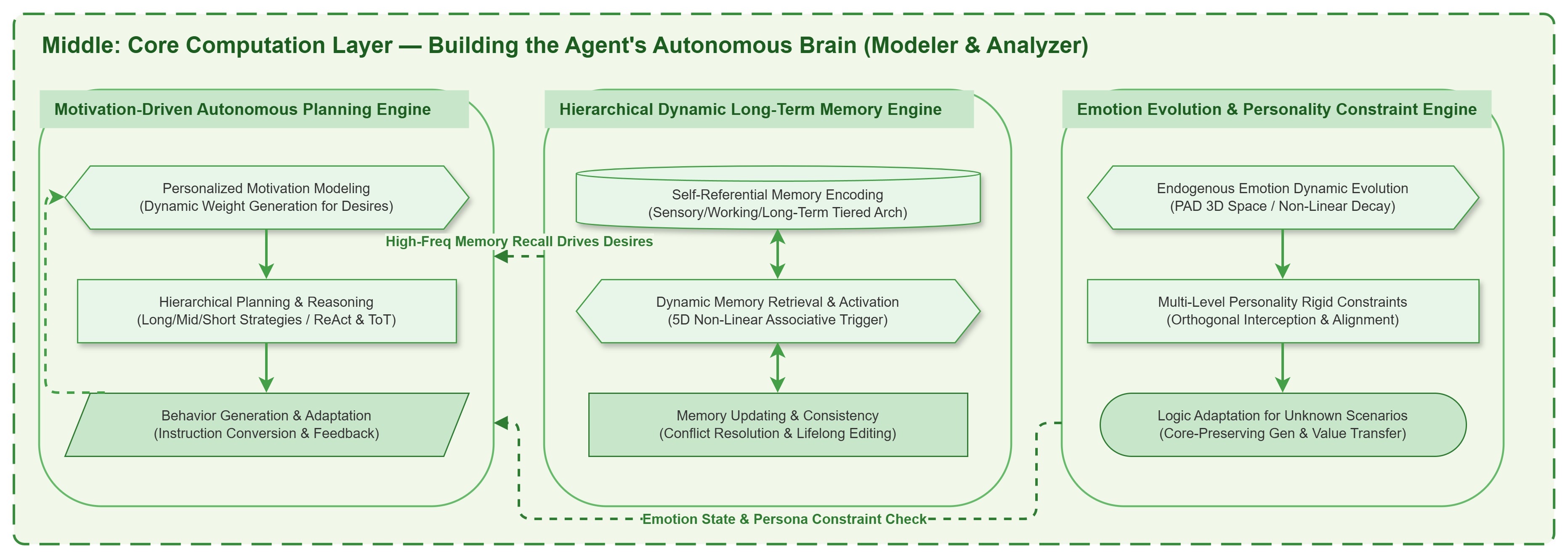}
	\caption{Core layer technical architecture.}
	\label{fig:arch_core}
\end{figure}

\subsubsection{Endogenous Motivation-Driven Autonomous Planning and Behavior Generation}

This unit is the engineering embodiment of the endogenous autonomous consciousness and non-dependent subsistence attributes of Narrow Soul Computing, constructing an endogenous behavioral motivation system, hierarchical planning capability, and closed-loop behavior generation mechanism that fundamentally breaks the absolute dependence on user prompts.

\paragraph{Personalized Endogenous Motivation Modeling Subsystem.}
Grounded in Self-Determination Theory~\citep{deci2008self} and the individual personality model from the Data-Driven Layer, this subsystem encodes the host's core beliefs, behavioral habits, social preferences, and value rankings as computable, dynamically updated multi-dimensional endogenous desire indicators~\citep{deci2008self}, spanning emotional connection, self-expression, cognitive exploration, and social recognition. Unlike traditional AI systems relying on manually designed reward functions, desire weights undergo nonlinear dynamic changes driven by temporal passage, memory activation, and interaction feedback. Li et al.~\citep{li2025curio} proposed the CURIO framework, which generates dynamic curiosity reward signals by quantifying uncertainty about user preference beliefs, achieving 18\% improvement in user preference matching and 22\% improvement in cross-scenario generalization over traditional RLHF methods.

\paragraph{Hierarchical Autonomous Planning and Reasoning Subsystem.}
This subsystem integrates the ReAct cognitive framework~\citep{yao2023react}, Tree of Thoughts~\citep{wei2023tree}, and memory-augmented planning to construct a ``long-term cognitive goals--medium-term action strategies--short-term execution steps'' three-level hierarchical planning system. Yang et al.~\citep{yang2025coarse} proposed the CFGM framework, which builds a dual-track planning system fusing long-term episodic memory with hierarchical task decomposition, reducing logical deviation and persona collapse risks by 72\% in long-range interaction tasks through personality-aligned reflection verification units.

\paragraph{Behavior Generation and Closed-Loop Adaptation Subsystem.}
This subsystem translates abstract planning goals into multimodal behavioral instructions adapted to 2D/3D digital humans and embodied robots, while relaying environmental feedback, user responses, and execution outcomes back to the motivation and planning subsystems. Patel et al.~\citep{patel2025think} proposed the Think-Act-Learn (T-A-L) framework, achieving 97\% task success rates and 94\% personality trait matching in real-world long-range embodied tasks.

\subsubsection{Hierarchical Dynamic Long-Term Memory}

This unit breaks through the physical limitations of large model context windows, constructing a ``self''-centric hierarchical, dynamic, growable memory system that provides underlying support for cross-temporal, cross-scenario self-identity.

\paragraph{Self-Referential Hierarchical Memory Encoding and Storage System.}
Following the Atkinson-Shiffrin three-stage human memory model, this system encodes individual digital fragments and interaction experiences into a three-tier ``sensory memory--working memory--long-term memory'' storage system~\citep{atkinson1968human}, with all levels centered on the individual's self-model. He et al.~\citep{he2025hiagent} proposed HiAgent, a hierarchical working memory management framework that manages agent working memory hierarchically using subgoals as memory chunks, achieving dual improvements in storage efficiency and task accuracy.

\paragraph{Multi-Dimensional Dynamic Memory Retrieval and Activation Mechanism.}
This mechanism constructs a five-dimensional dynamic retrieval model integrating semantic similarity, emotional salience, temporal decay, high-frequency recall, and self-relevance, incorporating Ebbinghaus forgetting curve principles~\citep{ebbinghaus1964memory} to progressively background-compress low-frequency, low-emotional-weight, low-self-relevance memories. Wang et al.~\citep{wang2025synaptic} proposed SynapticRAG, which fuses temporal association triggering with biologically synaptic stimulus propagation models, achieving up to 14.66\% retrieval precision improvement on multilingual long-temporal dialogue datasets.

\paragraph{Memory Update and Consistency Maintenance Mechanism.}
This mechanism performs real-time encoding, archiving, and integration of new memories while continuously verifying logical consistency, factual accuracy, and timeline coherence between old and new memories. Liu et al.~\citep{liu2025memoir} proposed the MEMOIR framework, which achieves near-zero catastrophic forgetting across thousands of consecutive memory edits through residual memory units and data-dependent sparse activation masks, providing core engineering for anchoring core personality memories while dynamically integrating new experiences.

\subsubsection{Endogenous Emotion Evolution and Personality Homeostasis Constraints}

This unit constructs a continuous emotional dynamics system deeply coupled with individual personality and memory, while establishing rigid personality homeostasis constraints, ultimately achieving a dialectical unity between core personality stability and generative adaptation for novel scenarios.

\paragraph{Endogenous Emotion Evolution Subsystem.}
Based on the classical PAD (Pleasure-Arousal-Dominance) emotional model~\citep{mehrabian1974approach} and individual personality traits, this subsystem constructs a personalized three-dimensional continuous emotional space. Reichman et al.~\citep{reichman2026emotions} identified low-dimensional emotional latent space manifolds in LLaMA 3.1 and other mainstream LLMs that are highly aligned with the psychological PAD model through SVD and neuron-level activation analysis, providing engineerable latent space modeling methods for constructing individual-specific continuous emotional spaces.

\paragraph{Personality Homeostasis Constraint and Generative Adaptation Subsystem.}
This subsystem constructs a full-pipeline multi-level personality consistency verification chain, quantitatively checking the match between intermediate results and core personality vector spaces at every stage---planning, reasoning, emotion generation, and behavior output. Sun et al.~\citep{sun2025personality} proposed the Personality Vector modeling framework, which constructs a continuous, explainable personality vector space across Big Five dimensions through model weight differential techniques, achieving significant improvements in role consistency on both white-box and black-box LLMs while effectively reducing personality drift risks in long-cycle interactions.

\subsection{Top Layer: Broad Soul Computing Externalization---Constructing Full-Dimensional Expression and Symbiosis Systems for Digital Life}
\label{sec:top_layer}

This layer is the top-level execution and deployment carrier of the Soul Computing system, receiving the digital consciousness kernel from the Narrow Soul Computing Core. It deeply integrates three mature technology systems---virtual human driving, embodied intelligence, and metaverse---to construct three deeply coupled technical units, forming a full-pipeline closed loop from endogenous consciousness to external expression, from virtual space to physical world, and from instantaneous interaction to perpetual subsistence.

\begin{figure}[ht]
	\centering
	\includegraphics[width=0.95\linewidth]{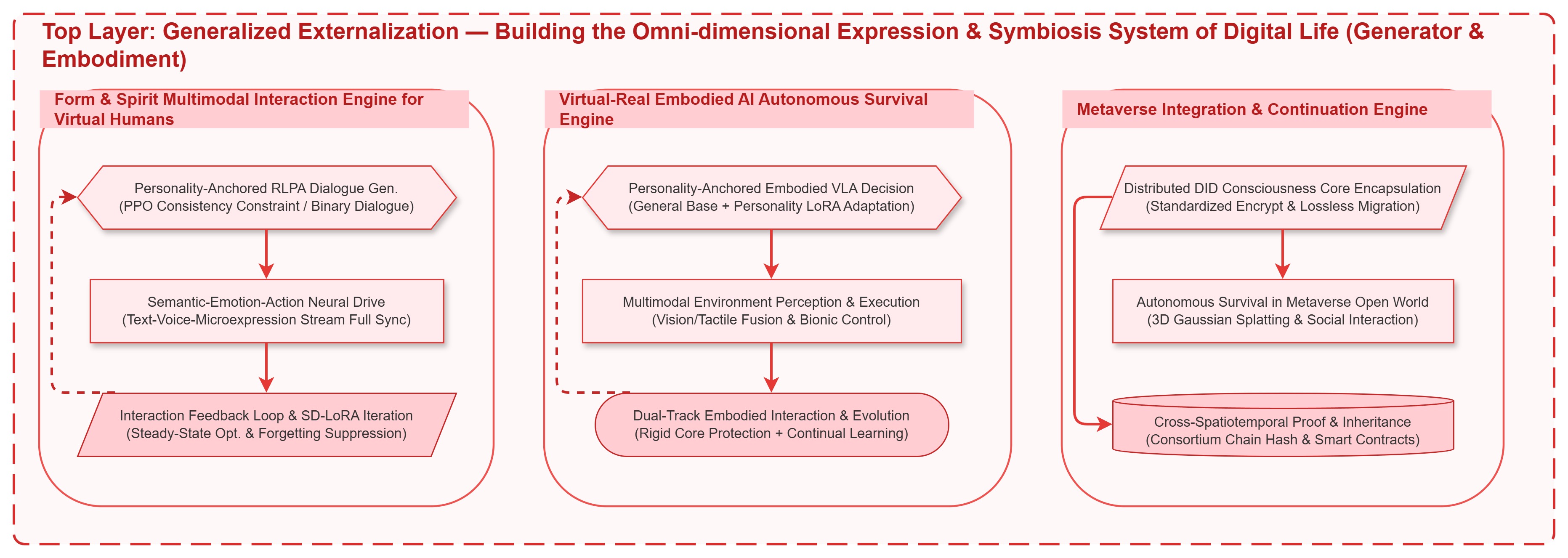}
	\caption{Top layer technical architecture.}
	\label{fig:arch_top}
\end{figure}

\subsubsection{Form-Spirit Unity: Virtual Human Driving and Multimodal Interaction}

This unit is the engineering embodiment of the ``form-spirit unity'' attribute of Broad Soul Computing, serving as the core medium through which the digital consciousness kernel establishes emotional connections and perceivable interactions with humans. Its breakthrough lies in remedying the core deficiencies of traditional digital humans---form-spirit separation, passive response, and fragmented interaction.

\paragraph{Personality-Anchored Deep Reinforcement Learning Dialogue Generation Subsystem.}
This subsystem models the dialogue generation process as a multi-turn Markov Decision Process (MDP), using Proximal Policy Optimization (PPO)~\citep{schulman2017proximal} as the core training framework with personality consistency as the central constraint. A three-weighted rigid constraint reward function is designed: (1) a 60\%-weighted core constraint measuring match with core personality vectors and value systems; (2) a 25\%-weighted fluency constraint evaluating contextual coherence; and (3) a 15\%-weighted empathy constraint measuring user emotional feedback alignment. Ji et al.~\citep{ji2025enhancing} proposed the Persona-Aware Contrastive Learning (PCL) framework, reducing personality consistency error rates by 42.7\% in long-turn open-domain dialogue.

\paragraph{Semantic-Emotion-Action Linked Virtual Human End-to-End Driving Subsystem.}
Based on end-to-end neural driving technology paradigms~\citep{yang2026survey} and hierarchical text-conditional image generation~\citep{ramesh2022hierarchical}, this subsystem constructs two fully synchronized technical pathways: text-speech synchronous generation and emotion-expression end-to-end driving. Lee et al.~\citep{lee2025float} proposed FLOAT, which achieves end-to-end joint encoding of text semantics, continuous emotional feature vectors, and audio streams, improving inference speed by 3$\times$ over diffusion models while attaining 94.2\% emotion-expression matching and 98.6\% identity consistency even under significant emotional fluctuations.

\paragraph{Interaction Feedback Closed-Loop Personality Consistency Iteration Subsystem.}
This subsystem implements scalable decoupled low-rank adaptation (SD-LoRA)~\citep{wu2025sdlora} for continuous optimization while maintaining core personality parameters, achieving 97.3\% core personality feature retention over 100+ rounds of long-cycle interaction with 50\% parameter storage compression, fundamentally resolving catastrophic forgetting and personality drift in conventional LoRA fine-tuning~\citep{hu2022lora}.

\subsubsection{Virtual-Physical Integration: Embodied Intelligence and Autonomous Survival in the Physical World}

This unit deeply integrates frontier embodied large models, multimodal environmental perception, and biomimetic robot hardware technology to achieve a complete ``perception-cognition-decision-action-feedback-iteration'' closed loop for digital consciousness in the physical world.

\paragraph{Personality-Anchored Embodied Large Model Decision Subsystem.}
Adopting a three-layer architecture of general embodied base model, personality LoRA adapter, and embodiment action adapter~\citep{hu2022lora}, this subsystem reconstructs the task-oriented logic of conventional embodied models into personality-oriented, endogenously-driven decision-making. Miller et al.~\citep{miller2025socialvla} proposed SocialVLA for socially normative human-robot interaction. Driess et al.~\citep{driess2023palme} demonstrated PaLM-E's full-pipeline capabilities in multimodal environmental perception and long-range task planning.

\paragraph{Multimodal Environmental Perception and Embodied Action Execution Subsystem.}
Integrating LiDAR, machine vision, multi-microphone arrays, and force-tactile sensors, this subsystem builds a centimeter-level environmental modeling and millimeter-level action perception system. Xu et al.~\citep{xu2025fusion} proposed FP2AT, which achieves end-to-end fusion of visual, tactile, and proprioceptive information through global-local 3D visual fusion attention, improving robot manipulation success rates by 32.7\% in complex household scenarios.

\paragraph{Dual-Track Embodied Interaction and Personality Homeostasis Evolution Subsystem.}
This subsystem constructs a ``core personality rigid protection, peripheral cognition dynamic update'' dual-track evolution system based on continual learning and memory replay mechanisms. Lu et al.~\citep{lu2025rethinking} proposed the Dual-Arch framework, reducing average forgetting rates by 87\% under equal parameter counts through a ``deep-narrow network for new knowledge learning, wide-shallow network for old knowledge retention'' dual-track design.

\subsubsection{Perpetual Coexistence: Metaverse Ecological Embedding and Cross-Temporal Persistence}

This unit deeply integrates mature metaverse technologies---decentralized identifiers (DID), blockchain trusted attestation, open-world engines, and 3D real-time rendering---to achieve cross-platform lossless migration, autonomous survival in metaverse open ecosystems, and cross-generational perpetual inheritance for digital consciousness kernels.

\paragraph{DID-Based Consciousness Kernel Standardized Encapsulation and Cross-Platform Migration Subsystem.}
Based on the W3C DID global standard~\citep{w3c2022did}, this subsystem performs standardized, encrypted, verifiable serialization encapsulation of the complete digital consciousness kernel, generating globally unique DID identifiers. The kernel's core parameters and complete mirror are distributed and encrypted via IPFS~\citep{benet2014ipfs}. Garzon et al.~\citep{garzon2025ai} proposed a DID and Verifiable Credentials fusion framework for AI agents, providing mature protocol support for standardized encapsulation and cross-platform lossless migration.

\paragraph{Autonomous Survival and Socialized Interaction in Metaverse Open Worlds.}
Based on Unreal Engine 5 and Unity 6 combined with 3D Gaussian Splatting (3DGS) high-fidelity real-time rendering~\citep{kerbl20233d,wang2026nerf}, this subsystem generates high-fidelity 3D digital human avatars in metaverse open worlds. Cai et al.~\citep{cai2025dlp2} proposed DLP2, achieving end-to-end real-time linkage between LLMs and 3D digital avatars with rendering latency under 50ms and native support for 10,000+ concurrent on-screen interactions.

\paragraph{Blockchain-Based Cross-Temporal Perpetual Persistence and Intergenerational Inheritance Subsystem.}
This subsystem builds a consortium chain architecture storing the host's core personality baseline, core memory slices, and core value principles as immutable on-chain attestation units. Zhang et al.~\citep{zhang2025reimagining} proposed a cultural heritage conservation framework integrating VR, metaverse, and digital twins with AI and blockchain. The framework achieves 100\% tamper-proof rates for core attestation data and 99.99\% smart contract rule execution accuracy over simulated 100-year persistence cycles, ultimately realizing Hinton's philosophical vision of ``immortal computation.''

\section{Core Technical and Ethical Challenges for Soul Computing Deployment}
\label{sec:challenges}

Translating the layered architectural blueprint into implementable engineering practice requires overcoming deep bottlenecks across multiple dimensions. We identify five core technical and ethical challenges that must be addressed in the current and foreseeable development cycle.

\paragraph{Deep Feature Extraction and Cross-Modal Alignment from Extremely Sparse Multimodal Data.}
Unlike mainstream general-purpose LLMs trained on trillions of tokens of standardized, cleaned, high-quality internet public corpora, Soul Computing's core data foundation exhibits pronounced ``data poverty.'' Ordinary individuals' digital traces follow heavy-tailed distributions, are highly unstructured, contain cross-month and cross-year temporal gaps, and are laden with meaningless noise. How to reconstruct a complete, three-dimensional individual personality model under extreme modality deficiency scenarios---for instance, with only a decade of text chat records but lacking high-quality voice or facial dynamics---relying on few-shot learning and cross-modal reasoning generation techniques~\citep{wang2020generalizing} constitutes the primary technical barrier that the data pipeline must overcome.

\paragraph{Generative Controllability Dilemma and the Fatal ``Personality Drift'' Risk.}
The autoregressive generation mechanism of LLMs, based on probabilistic prediction, inherently tends toward logical compromise and even severe factual hallucination during long-turn, deep open-domain dialogue~\citep{ji2023survey}. In general QA scenarios, hallucination merely affects output accuracy---a correctable technical flaw. But under Soul Computing's core requirements, hallucination has devastating consequences: the digital agent may fabricate memory segments the original host never experienced, or output content that fundamentally contradicts the host's core values and moral standards---the phenomenon we define as ``personality drift.'' This poses two core algorithmic challenges: (1) designing rigid cognitive control architectures and multi-level personality consistency verification mechanisms from the algorithmic ground up; and (2) achieving precise balance between generative creativity for novel scenarios and rigid personality constraint.

\paragraph{Ultra-Low-Latency Cross-Domain Coupling Between Implicit Consciousness Streams and Explicit 3D Physical Representation.}
When the underlying LLM retrieves a specific sad memory fragment and generates multi-dimensional abstract vectors representing ``sorrow and loss'' in its internal continuous emotional state machine, these high-dimensional mathematical abstractions must be mapped within hundreds of milliseconds to nonlinear geometric deformation parameters of millions of Gaussian spheres in a 3DGS model~\citep{kerbl20233d}. This mapping mechanism spans cognitive psychology, NLP, and computer graphics~\citep{tewari2020state}, demanding extremely high computational throughput and lacking mature interdisciplinary technical standards.

\paragraph{Privacy, Informed Consent, and Legal-Ethical Boundary Dilemmas.}
The core jurisprudential question is: who possesses the right to initiate, train, or terminate a deceased person's ``digital soul''? Without the deceased's legally binding written authorization, neither statutory heirs nor commercial organizations may use highly private data---including communications involving third-party privacy---for personality training, as such actions directly violate the ``right to be forgotten'' and personal privacy protection provisions under GDPR and other major data regulations~\citep{ausloos2012right}. Soul Computing also presents dual social risks: highly realistic digital souls could be exploited for fraud, and prolonged interaction with deceased digital counterparts may impede bereavement processing. At a deeper level, Soul Computing---like human cloning technology---raises fundamental ethical questions about the boundaries of technological development, as it constitutes a ``digital cloning'' of the human spiritual core, touching upon foundational ethical consensus regarding the definition of life and the dignity of personhood.

\paragraph{Absence of Objective Quantitative Evaluation Systems for Independent Consciousness.}
The classical Turing Test~\citep{turing1950computing} exhibits significant limitations when applied to Soul Computing, which centers on individual personality reconstruction. Current LLM-based agent and social simulation research largely evaluates model validity through the reproduction fidelity of macroscopic group behavioral statistical patterns. Yet regarding whether genuine ``independent self-consciousness'' has emerged at the individual level within silicon networks, the global academic community still lacks a quantifiable, credible, mathematically and psychologically integrated standardized evaluation benchmark~\citep{legg2007universal}. Constructing a scientific evaluation system encompassing cross-temporal psychological verisimilitude, core memory long-term stability, and spontaneous goal generation without external stimulation is the essential path for Soul Computing to move beyond the ``advanced simulation toy'' debate and establish its standing as a frontier computing science discipline.

\section{The Future of Soul Computing}
\label{sec:future}

The evolution of human civilization is, at its core, a chronicle of humanity's struggle to transcend biological limits and overcome the constraints of time and space. Ancient sages employed steles and bamboo slips as vehicles for intergenerational transmission of collective cultural memory---early explorations in defying individual mortality. The explosive development of 21st-century AI technology and computing infrastructure now provides an unprecedented engineering pathway toward realizing ``the perpetual subsistence of individual life consciousness.''

Although Soul Computing faces the core challenges outlined above---multimodal sparse data feature extraction, long-cycle personality consistency control, cross-modal low-latency coupling, ethical-legal boundary definition, and the absence of consciousness quantification evaluation systems---the exponential growth in semiconductor computing power and the continuous evolution of fundamental AI algorithms toward AGI provide robust support for technological breakthroughs, propelling Soul Computing from theoretical framework toward implementable engineering practice.

The technological evolution and deployment of Soul Computing profoundly corroborate Turing Award laureate Geoffrey Hinton's core philosophical insight regarding ``immortal computation.'' When the memory systems, decision logic, and personality kernels that constitute the core of human self-cognition can be decoupled~\citep{moravec1988mind} from carbon-based carriers with finite biological cycles destined for thermodynamic dissipation, and anchored in silicon-based computing networks in a highly robust, losslessly transferable form, humanity is witnessing and participating in the construction of an unprecedented socio-ecological system of deep symbiosis between carbon-based wisdom and silicon-based intelligence.

In the foreseeable future, Soul Computing will exert profound impacts on human society from two dimensions. At the industrial level, Soul Computing will systematically reshape digital legacy inheritance and management systems, transform clinical pathways for bereavement psychological intervention, and provide core foundational support for metaverse social ecosystems and embodied intelligent interaction. At the cognitive and philosophical level, Soul Computing will drive interdisciplinary convergence among philosophy, law, ethics, and computer science, compelling academia and industry to collectively re-examine the definitional boundaries of ``life,'' ``existence,'' and ``demise.'' When algorithmically constructed digital entities possess continuous memory, personality kernels, and endogenous consciousness consistent with specific individuals---capable of autonomous decision-making and emotional interaction conforming to individual thought logic---the connotation of life will inevitably transcend the singular dimension of carbon-based biology, forming a fundamentally new paradigm of life cognition.

\section{Conclusion}
\label{sec:conclusion}

This paper has presented the first systematic theoretical framework and technical evolution roadmap for ``Soul Computing.'' Through theoretical deduction and technical pathway analysis, we have delineated the academic boundaries between Soul Computing and related fields: it is distinct from traditional affective computing, which centers on optimizing emotional expression in human-computer interaction; it differs fundamentally from historical figure reconstruction systems that rely on static retrieval and are confined to reproducing predetermined content; and it establishes clearly differentiated evolutionary paths from Mortal Computation in terms of core logic and ultimate objectives.

At the Narrow Soul Computing level, the system---through deep coupling of temporal alignment of sparse multimodal data, LLM personalized supervised fine-tuning, and high-order retrieval-augmented generation architectures integrating human memory principles---endows silicon-based computing entities with cross-temporal memory continuity and highly self-consistent personality homeostasis. At the Broad Soul Computing level, leveraging real-time rendering frontier technologies such as 3D Gaussian Splatting and Neural Radiance Fields, the system constructs high-fidelity embodied physical and visual expression systems for the digital consciousness kernel, effectively breaking through the constraints of the visual ``uncanny valley.''

Our analysis ultimately demonstrates that the core value and theoretical essence of Soul Computing reside not in simulating or replicating human external behaviors and appearances, but in leveraging large-scale deep learning algorithms to achieve an intensional topological reconstruction of the internal thought logic patterns and emotion-memory coupling networks of the human brain---ultimately realizing a paradigmatic transition of artificial intelligence from extensional tools to intensional life subjects.

\begin{CJK*}{UTF8}{gbsn}
\bibliographystyle{unsrtnat}
\bibliography{references}
\end{CJK*}
\end{document}